\documentclass[preprint,showpacs,showkeys,preprintnumbers,amsmath,amssymb]{revtex4}

\usepackage{graphicx}
\usepackage{dcolumn}
\usepackage{bm}
\usepackage{epsfig} 
\usepackage{csquotes}
\usepackage{mathtools}
\usepackage{algorithm}
\usepackage[noend]{algpseudocode}
\usepackage[lofdepth,lotdepth]{subfig}
\begin{document}

\preprint{APS/123-QED}

\title{Learning in Competitive Network with Haeusslers Equation adapted using FIREFLY algorithm  }

\author{N. Joshi}
\email{njoshi@fias.uni-frankfurt.de}


\affiliation{%
Institute for Applied Physics, Goethe University, \\
60438 Frankfurt, Germany 
}

\date{}

\begin{abstract}

Many of the competitive neural network consists of spatially arranged neurons.
The weigh matrix that connects cells represents local excitation and long-range inhibition.
They are known as soft-winner-take-all networks and shown to exhibit desirable information-processing.
The local excitatory connections are many times predefined hand-wired based depending on  spatial arrangement which is chosen using the previous knowledge of data. 
Here we present learning in recurrent network through Haeusslers equation and modified wiring scheme based on biologically based Firefly algorithm.
Following results show learning in such network  from input patterns without hand-wiring with fixed topology.

\end{abstract}



\maketitle

\section{\label{sec:level1}Introduction\protect\\
 }

The concept of competitive learning and self organization in neural networks was first introduced by Malsburg (1973)  successfully demonstrated for retinotopy \cite{Mals, Haeuss}.
Later developed many of the competitive neural network consists of spatially arranged neurons.
The weigh matrix that connects cells represents local excitation and long-range inhibition.
The "soft-winner-take-all" dynamics is commonly used in such networks.
The "soft-winner-take-all" dynamics consists of local group of  the winners rather than single winner as in "hard " winner in Winner-Take-All dynamics \cite{WTA, SWTA}.
The Self-Organizing Maps is another example that  uses a similar mechanism \cite{SOM}.
In case of WTA networks the network topology is mostly predefined and spatially fixed. 
The soft-WTA mechanism can be implemented by mexican hat fuction or also called Ricker function.
With one winning neuron and immidiate neighbours connected with positive weights decreasing with the distance and negative weights on far distant neurons.

A FIREFLY algorithm was developed in order to simulate soft-WTA network with adaptive network topology.
The dynamics of frequency synchronization in fireflies was described by Hanson in (1978).
Here we simulate the spatial development of neurons modelled as firelies that can be used to form a population of inhibitory and excitory neurons \cite{firefly, firefly_2, firefly_3}.
The disadvantege of this system would be one additional computational step and the memory usage.
But on the other hand the efficiency of recall and  signal reconstruction using this algorith is much better.
In the following sections we describe the learning mechanism using the Haeussler's equation and the improvement achieved using FIREFLY algorithm.

\section{\label{sec:level1}Haeusslers Equation\protect\\
 }

The Haeussler's equation can be written in the following  form:


\begin{equation}
\begin{split}
f(w_{ij}) =    \alpha \left( 1 - N w_{ij} \right) + \\
& + \beta w_{ij}(T^{S^\mu}_{ij} -  \sum_{j'}w_{ij'} T^{S^\mu}_{ij'})  \\
&  : for ~i,j,j' \in N .
 \end{split}
\end{equation}

with additional saturation condition:

\begin{equation}
\dot{w}_{ij} = G_{sat} \times f(w_{ij})
 \end{equation}


Here, the function $g_v(x)$ enforces saturation of the synaptic weights at the value v such that
$G_{sat}(x) = 1~for~x \leq v, = 0 ~for ~x > v$. 
The first term by itself is trying to pull the weights towards the value $1/N$, whereas the second term by itself would let $\sum_{j'}w_{ij'}$ eventually converge towards $1$, so that the sum in the second term could be interpreted as a weighted average: $\bar{T}^{S^\mu} := \sum_{j'}w_{ij'}T^{S^\mu}_{ij'}$ .
The second term, then, would let those weights that have above-average cooperative help $T^{S^\mu}$ grow at the expense of the below average cooperation weights.
The weights will thus favor the higher cooperation terms, increasing $\bar{T}^{S^\mu}$, letting more weights fall below-average cooperation and thus to start decaying, until finally only  weight(s) with maximal cooperation will stabilize. 
If $\alpha$ is positive, the system is driven towards a compromise between the first and the second term of eq .
The initial state of $W$ should be random, with the probability or strength of a connection between cells $i$ and $j$ falling off monotonously with the distance between the two in a two-dimensional plane. 
For detailed explaination on equation see Appendix.

\section{Response of the equation}

\begin{figure}[h]
\centering
\subfloat[][]{
\includegraphics[width=4cm]{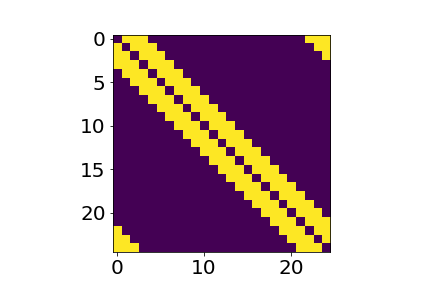}
\includegraphics[width=4cm]{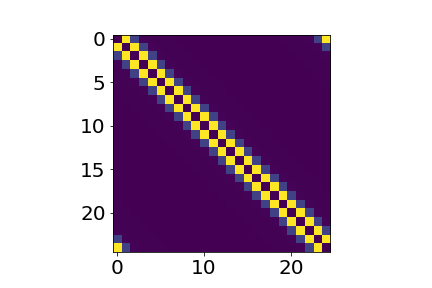}
\label{a}
}
\qquad
\subfloat[][]{
\includegraphics[width=4cm]{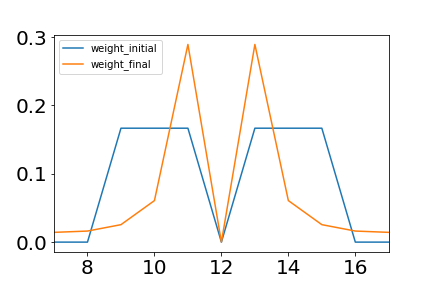}
\label{b}
}
\qquad
\caption{Evolution of weight matrix $w_{ij}$. $N= 25$ with $3$ lateral connections.(a): Initial and final $w_{ij}$ matrix,(b): Initial  and final weight vector for $i=12$.}
\label{Evolve_1}
\end{figure}

Consider $N = 25 $ number of neurons with two lateral connections.
Fig. \ref{Evolve_1} shows evolution weight matrix and weight vector for $i=12$ under modified equation.
Following conditions are assumed:
\begin{itemize}
\item $w_{ij} = 0$ for $i =j$, no self connections
\item $ \sum w_{i} =1.0$
\end{itemize}

Thus if we have two neighbouring connections, the initial condition would be, 

\begin{itemize}
\item $w_{ii} = 0$ 
\item $  w_{i, i-1} =   w_{i, i-2} =  w_{i, i+1}  w_{i, i+2} = 0.25$
\end{itemize}

In the example $3$ neighbours are assumed. 
To avoid any boundary effects a periodic condition is assumed.
Fig. \ref{Evolve_1} shows immediate neighbour strengthening and others lowering the weights.

\section{\label{sec:level1}Input Patterns\protect\\
 }

\begin{figure}[h]
\centering
\includegraphics[width=8cm]{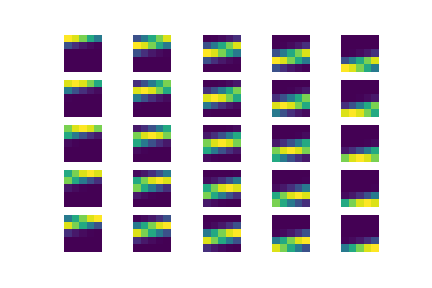}
\includegraphics[width=8cm]{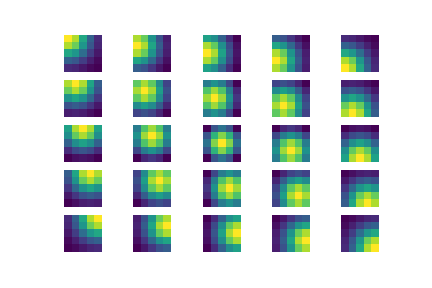}
\caption{ $1D$ and $2D$ input patterns reshaped into $2D$. Example with $N=25$ neurons.}
\label{input_1}
\end{figure}

Two types of input patterns were used, one dimensional and two dimensional defined through  Gaussian function.
Fig. \ref{input_1} shows $1D$ and $2D$ input in $5 \times 5$ square form.

In one dimensional case the input pattern given as,
\begin{equation}
G = \exp {  \left( \frac{1}{2} \left( \frac{x- \mu}{\sigma}\right)^2 \right) },
\end{equation}
where $\mu$ is the central value varied over the number of neurons.
Similarly in $2D$ case the $\mu_x$ and $\mu_y$ varied with $i,j \in N$ where $N$ is total number of neurons.
\begin{equation}
G = \exp {  \left( \frac{1}{2} \left( \frac{x- \mu_x}{\sigma_x}\right)^2 + \frac{1}{2} \left( \frac{y- \mu_y}{\sigma_y}\right)^2 \right) }
\end{equation}

The input vectors are normalized and  generated at random for tranning and fed to the network.

\section{\label{sec:level1}FIREFLY Algorithm\protect\\
 }

In earlier discussed cases 2D grid is defined.
To modify this scenario a population based firefly algorithm is developed.
The female fireflies attracted towards dominant male flies depending on brightness.
The frequency driven algorithm was developed by Hanson (1978).
The excitory neurons and inhibitory neurons are attracted to active cell depending on the activity level or brightness.
Thus forming a population of neurons which is used to define a weight matraix.
The movement of neurons is governed by equations:

\begin{equation}
B(r) = b e^{\gamma r^2}
\end{equation}

where $B$ is the brightness of a 'male firefly' to which the 'female' fireflies are attracted.
The frequency dependence is ignored in this case.

\begin{equation}
w_i = w_i + b e^{\gamma r^2} (w_j-w_i)+ \eta \left(  rand - 1/2 \right)
\end{equation}

After feeding each pattern to the network the population of fireflies is redistributed thus redefining the distribution of of inhibitory and excitory neurons.
The weight matrix then subjected to the Haeuslers equation thus reducing number of excitory neurons and inhibitory neurons those represent particular pattern and remain active.

\begin{algorithm}
\caption{Psuedo code Firefly}\label{euclid}
\begin{algorithmic}[1]
\Procedure{FIREFLY}{}
\State Generate initial population $X = {x1, x2, ...x_n}$
\State Compute brightness using objective function 

		$B = {B_1,B_2, ...B_n} = {f(x_1),f(x_2)...f(x_n) }$ 

\State Move fireflies $i$ and $j$ recalculate positions according to attractiveness
\State Calculate minimum distance condition
\EndProcedure
\end{algorithmic}
\end{algorithm}

\begin{algorithm}
\caption{Psuedo code Weight Calculation}\label{euclid}
\begin{algorithmic}[1]
\Procedure{Weights}{}
\State Generate $W$ matrix
\State Generate initial population $X = {x1, x2, ...x_n}$
\State Run FIREFLY CODE
\State Read pattern and set activity
\State Calculate weight matrix
\State Run weight learning algorithm
\State Get new pattern
\EndProcedure
\end{algorithmic}
\end{algorithm}

\section {Results}
\label {results}

For demonstration purpose low number of  neurons are used.
Previously described 2D input patterns are fed.
The input pattern can either be presented after convergence or 

\begin{figure}[!h]
\vspace{0.5cm}
\begin{center}
\includegraphics[ width = 4 cm]{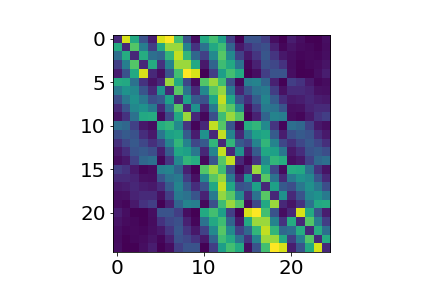}
\includegraphics[ width = 4 cm]{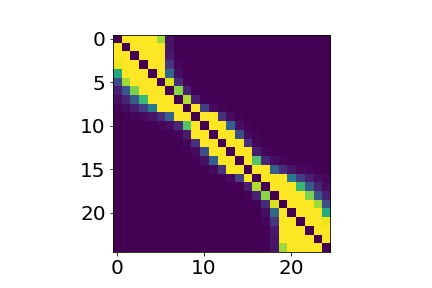}
\end{center}
\caption{Network of  $N=25$, fed with random 2D pattern.  (a): Input patterns, (b):  Output without FIREFLY,  . }
\label{D2_pat_3}
\end{figure}

The learning algorithm can either run when a new input is first presented or only
after the network has converged.
The scenario of running learning algorithm  at stimulus onset is more  biophysiology compatible.
Fig. \ref{D2_pat_3} shows the learnt weight matrix for two dimensional case.

In the earlier examples the periodic boundary condition was not applied. 
Hence one observes assymetry at $i =0~ and ~ i=25$.

\begin{figure}[!h]
\vspace{0.5cm}
\begin{center}
\includegraphics[ width = 4 cm]{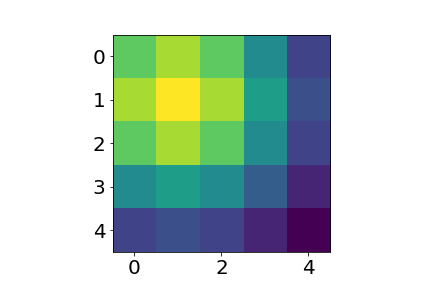}
\includegraphics[ width = 4 cm]{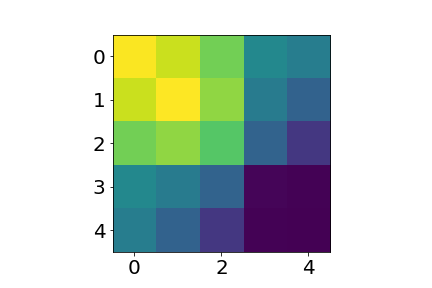}
\includegraphics[ width = 4 cm]{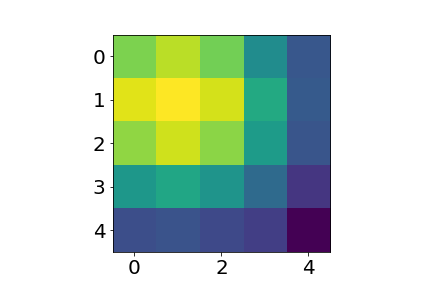}
\includegraphics[ width = 4 cm]{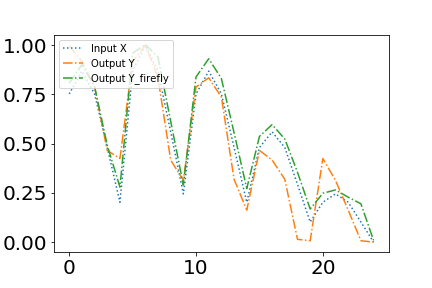}
\end{center}
\caption{Network of  $N=25$, fed with random 2D pattern.  (a): Input patterns, (b):  Output without FIREFLY, (c): Output with FIREFLY routine, (d): Input and output signals, with and without FIREFLY . }
\label{D2_pat_4}
\end{figure}

Fig. \ref{D2_pat_4} compares the recalled pattern from memory with and without FIREFLY subroutine. 
One clearly observes the the recalled pattern carries matches with the input pattern more closely when FIREFLY is applied.

\begin{figure}[!h]
\centering
\includegraphics[width=4.cm]{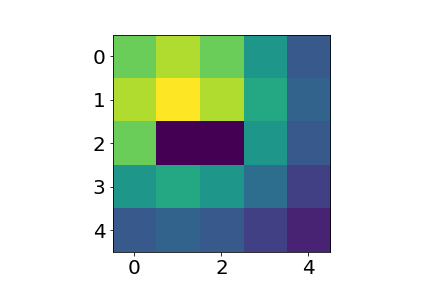}
\includegraphics[width=4.cm]{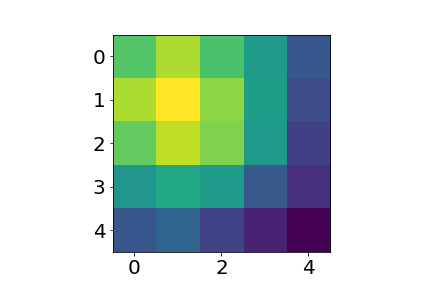}
\includegraphics[width=4.5cm]{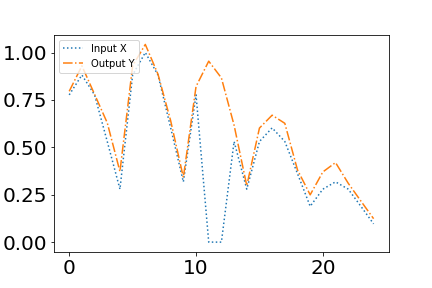}
\caption{Network of  $N=25$, fed with random  2D pattern. (a): Input pattern consisting with noise, (b): Recovered signal by trained network with FIREFLY subroutine, (c): Input and output signal.}
\label{noise_1}
\end{figure}

This type of  recurrent competitive networks is also able to complete partial patterns
Fig. \ref{noise_1} shows the signal recall when partially distorted pattern is presented.
After training the network  the excitatory connections are  strengtened between units whose activity is highly correlated .
This provides solution to the missing information.

\begin{figure}[!h]
\centering
\includegraphics[width=4.cm]{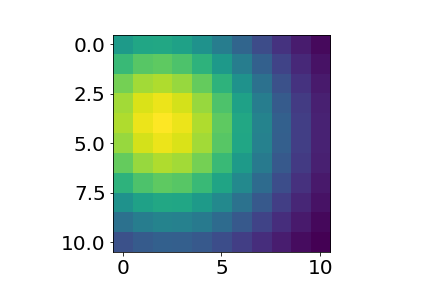}
\includegraphics[width=4.cm]{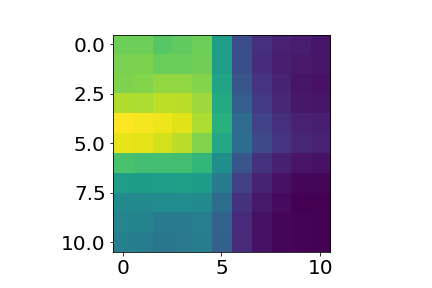}
\caption{Network of  $N=121$, fed with random  2D pattern. (a): Two Input pattern fused  with noise, (b): Recovered signal by trained network with FIREFLY subroutine}
\label{noise_2}
\end{figure}

When two fused inputs are presented it treats the input patters as similar input as shown in Fig \ref{noise_2}.
Or if one of the two inputs is weaker than the other it captures the difference between them.

\subsection* {An Example with Handwritten digits}
\label {example}
As an "real world" example we consider images of Handwritten digits, a dataset  available on internet for standard case.
A model image of each digit is prepared and fed to the network for learning.

\begin{figure}[!h]
\centering
\includegraphics[width=4.cm]{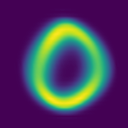}
\includegraphics[width=4.cm]{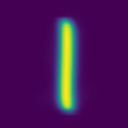}
\includegraphics[width=4.cm]{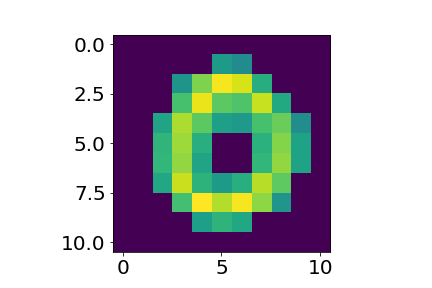}
\includegraphics[width=4.cm]{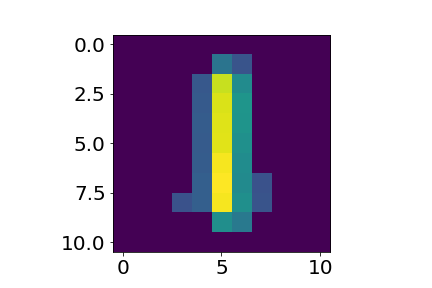}
\includegraphics[width=4.cm]{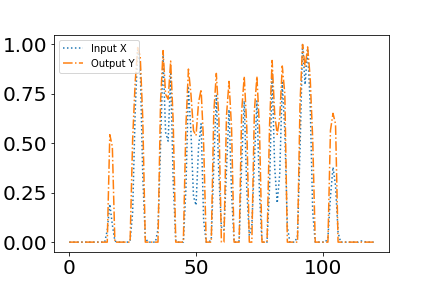}
\includegraphics[width=4.cm]{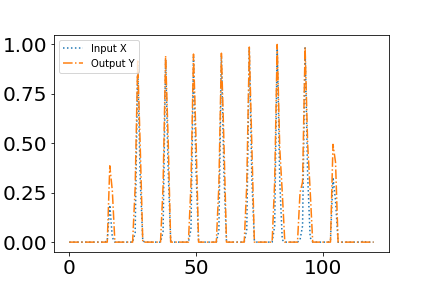}
\caption{The stored patterns and retrieved patterns using FIREFLY algorithm with two examples as digits. The figure on the bottom shows input and output signal from neurons.}
\label{numbers_1}
\end{figure}

Then noisy image is fed to the network which then outputs the corresponding stored image. 
Fig. \ref{numbers_1} shows the stored patterns and retrieved patterns using FIREFLY algorithm.

\section {Discussion}
\label {Discussion}

Here we have shown that the biologically motivated FIREFLY algorithm can be used for  memory based neural network with competitive learning.
In this paper we have shown that the network has ability to act as associative memory,  retrieve the information and handle the noisy signal or recalling partial information.
One of the disadvange would be that it requires additional calculation time to simulate firefly algorithm.
But the spotaneous population based algorithm makes it topology independent network.
Further efforts involve optimization of number of excitatory and inhibitory cells for which we would be implementing Lokta Volterra equations.

\appendix*
\section{}
\label{A}
To describe an associative memory for fragments of image content.
Each fragment has the form of a 2D neural field with short-range connections. Fragments may tile a larger visual domain.
This visual domain may be $V1$, and the fragments correspond to texture patches.
After some training the collections of all fragments should represent a codebook for all texture patches occurring in the input with some statistical significance.
The visual domain is composed of so-called columns. Each column stands for small resolvable region of the image and contains a full set of texture neurons, e.g., neurons with receptive fields in the form of Gabor wavelets. 
A particular visual input into $V1$ selects a subset $S^\mu$ of all the neurons of $V1$.
$S^\mu$ is the set of indices of those neurons that are active in state $\mu$. A state is assumed to be active for a typical period of length $\tau_2$.
The goal of the associative memory is to extract from a long sequence of image inputs a set of states $S = \{ S^\mu,\mu = 1,...,M\}$.
We are restricting consideration to image patches of a certain diameter $R$.
Let us designate by $N$ resp.
$N$ the set and number of neurons contained in a patch. 
The whole image domain is composed of a tiling of such patches, and an infinite number of possible images can be represented combinatorially by each patch selecting one of its states.
Whereas the general idea of associative memory is that the states $S^\mu$ are stored by plastically strengthening all connections between the neurons that are simultaneously active in a state, we want the connections between neurons active within a patch's activity state to form a network pattern (a "net fragment") that is an attractor under network self-organization. 
In network self-organization the activity of the neurons addressed by the input fluctuates spontaneously, and connections are changed by Hebbian plasticity. 
This is a positive feedback loop, as excitatory connections generate signal fluctuations, which in turn strengthen connections. 
This instability is checked by a constraint limiting the weight sum of connections converging on a neuron.
Given enough time this loop runs into an attractor state that is stabilized by the balance between Hebbian plasticity and the weight sum constraint.
We thus assume that at a given time the input selectively activates neurons $i \in S^\mu$ with the help of sources $s_i(t), for ~ i \in S^\mu$ (whereas $s_i(t) = 0$ for $i \not \in  S^\mu$).
These source terms fluctuate with a time constant $\tau_1 < \tau_2$ and have correlations $< s_i s_j >t= \delta_{ij}$, where the temporal average $< ... >_t$ is taken over the time scale $\tau_2$.
We assume further that the propagation of signal deviations from the average is very fast, at a time scale $\tau_0 << \tau_1$, so that we get equilibrium signal deviations $c_i(t)$ changing at the same time scale as the sources $s_i(t)$. These equilibrium deviations, then, obey the equation.

\begin{equation}
c_i(t) = \sum_{j \in N} w_{ij} c_j + s_i(t)
\end{equation}
 
 or in vector notation 
 
\begin{equation}
\vec c(t) =  w \vec c  + \vec s (t)
\end{equation}
 
 which is solved by
 
\begin{equation}
\vec c(t) =  (1- w)^{-1}  \vec s (t)
\end{equation}

Expanding that matrix inverse as

\begin{equation}
  (1- w)^{-1}  = 1 + w + w^2 + w^3 + \dots
\end{equation}

and braking this series off after the third power we have an approximate inverse which we call $D$, so that $\vec {c}(t) = \vec{D}s(t)$.

With this in hand we can compute temporal correlations of signals

\begin{equation}
\begin{split}
 < c_i c_j >_t =<  \sum_{k \in S^\mu } D_{ik} s_k \cdot \sum_{k' \in S^\mu } D_{jk'} s_{k'} >_t  \\
 =  \sum_{k \in S^\mu } D_{ik} D_{kj} = : T^{S^\mu}_{ij} ,
\end{split}
 \end{equation}

where we have used

\begin{equation} 
< s_i s_j >_t =
\begin{dcases}
    1 , & \text{if } i = k \in S^mu \\
    0.              & \text{otherwise}
\end{dcases}
\end{equation}

 (If $D$ is symmetric, $T^{S^\mu} = D^2$.) This result is easily interpreted: Fluctuations are communicated from the source cell to the two cells directly or over one or two or three intermediate connections.
 
 These correlations are to be used to drive Hebbian plasticity. Using the H\"{a}ussler equation as model we formulate it as:
 
  \begin{equation}
\begin{split}
 \dot{w}_{ij} =g_v(w_{ij}) \left( f_{ij}(W) - w_{ij} B_ij(f(W)) \right)  \\
	:for ~ i\neq j ~i,j  \in N ,
\end{split}
 \end{equation}

 which contains the cooperation term
 
 \begin{equation}
f_{ij}(W) = \alpha + w_{ij}T^{S^\mu}_{ij} ,
\end{equation}
 
 and the \emph{competition term}
  \begin{equation}
B_ij(f(W)) = \sum_{j'} f(w_{ij'}) .
\end{equation}

 Here, $\alpha$ is an unspecific synaptic growth rate.
 Eq. (6) can be re-written as

\begin{equation}
\begin{split}
f(w_{ij}) =    \alpha \left( 1 - N w_{ij} \right) + \\
& + \beta w_{ij}(T^{S^\mu}_{ij} -  \sum_{j'}w_{ij'} T^{S^\mu}_{ij'})  \\
&  : for ~i,j,j' \in N .
 \end{split}
\end{equation}


\begin{thebibliography}{99}

\bibitem{Mals} Chr. von der Malsburg, "Self-organization of orientation sensitive cells in the striate cortex"
\bibitem{Haeuss} A. F. Haeussler and Christoph von der Malsburg, "Development of retinotopic projections - an analytical treatment", Journal of Theoretical Neurobiology, 2:47-73, 1983. 
\bibitem{WTA} R. Douglas and K. Martin, “Recurrent neuronal circuits in the neocortex", Current Biology,  vol. 17, no. 13, pp. 496–500, 2007.
\bibitem{SWTA} S. Amari, “Dynamics 6 of pattern formation in lateral-inhibition type neural fields,” Biological 8 Cybernetics, Jan. 1977.
\bibitem{SOM} T. Kohonen
\bibitem{firefly} Janmenjoy Nayak et al , " Anovel nature inspired firefly algorithm with higher order neural network: Performance analysis" EST, 2015
\bibitem{firefly_2} Surafel Luleseged Tilahun and Hong Choon Ong, "Modified Firefly Algorithm", Journal of Applied Mathematics, Volume 2012, Article ID 467631,2012
\bibitem{firefly_3}Iztok Fister et al, "A comprehensive review of firefly algorithms"

\end{thebibliography}
\end{document}